\documentclass{article}
\usepackage{spconf,amsmath,graphicx}
\usepackage{booktabs}
\usepackage{multirow}
\usepackage{hyperref} 

\title{LadderNet: Multi-path networks based on U-Net for medical image segmentation}
\name{Juntang Zhuang}
\address{Biomedical Engineering, Yale University, New Haven, CT, USA}
\begin{document}
%
\maketitle
\begin{abstract}
U-Net has been providing state-of-the-art performance in many medical image segmentation problems. Many modifications have been proposed for U-Net, such as attention U-Net, recurrent residual convolutional U-Net (R2-UNet), and U-Net with residual blocks or blocks with dense connections. However, all these modifications have an encoder-decoder structure with skip connections, and the number of paths for information flow is limited. We propose LadderNet in this paper, which can be viewed as a chain of multiple U-Nets. Instead of only one pair of encoder branch and decoder branch in U-Net, a LadderNet has multiple pairs of encoder-decoder branches, and has skip connections between every pair of adjacent decoder and decoder branches in each level. Inspired by the success of ResNet and R2-UNet, we use modified residual blocks where two convolutional layers in one block share the same weights. A LadderNet has more paths for information flow because of skip connections and residual blocks, and can be viewed as an ensemble of Fully Convolutional Networks (FCN). The equivalence to an ensemble of FCNs improves segmentation accuracy, while the shared weights within each residual block reduce parameter number. Semantic segmentation is essential for retinal disease detection. We tested LadderNet on two benchmark datasets for blood vessel segmentation in retinal images, and achieved superior performance over methods in the literature. The implementation is provided \url{https://github.com/juntang-zhuang/LadderNet} 
\end{abstract}
\begin{keywords}
LadderNet, segmentation
\end{keywords}
\section{Introduction}
\label{sec:intro}
Deep learning has achieved state-of-the-art performance in many computer vision tasks, such as image classification, semantic segmentation, object recognition, motion tracking and image captioning \cite{lecun2015deep}. Convolutional neural networks have become popular since the success of AlexNet \cite{krizhevsky2012imagenet} on the classification task for ImageNet dataset \cite{deng2009imagenet} mainly for the following reasons: first, large datasets and powerful computational resources are available nowadays; second, convolutional operation on an image is translation-invariant and enables weight sharing for feature extraction; third, the success of activation functions such as rectified linear unit (ReLU); fourth, efficient optimization algorithms such as stochastic gradient descent (SGD) and Adam optimizer. 
\par
Deep convolutional neural networks (CNN) have achieved near-radiologist performance in many semantic segmentation tasks in medical image analysis. Fully convolutional neural network (FCN) \cite{long2015fully} have succeeded in semantic segmentation on Pascal VOC dataset, and U-Net \cite{ronneberger2015u} achieved the top accuracy in the segmentation of neuronal structures in electron microscopic stacks. Other variants of CNN achieve state-of-the-art performance on benchmark semantic segmentation tasks, such as PSPNet \cite{zhao2017pyramid} and DeepLab \cite{chen2018deeplab}. Among all these variants, U-Net is the most widely used structure in medical image analysis, mainly because the encoder-decoder structure with skip connections allows efficient information flow, and performs well when sufficiently large datasets are not available. 
\par
Many variants of U-Net have been proposed: Alom et al. proposed to use recurrent convolution in U-Net \cite{alom2018recurrent}; Ozan et al. proposed to use attention module in U-Net to determine where to look at; Simon et al. proposed Tiramisu \cite{jegou2017one}, where the convolutional layers in a U-Net are replaced with dense blocks. However, all these variants still fall into the encoder-decoder structure, where the number of paths for information flow is limited. We propose LadderNet, a convolutional network for semantic segmentation with more paths for information flow. We demonstrate that LadderNet can be viewed as an ensemble of FCNs, and validate its superior performance on blood vessel segmentation task in retinal images.  

\begin{figure*}[!htb]
    \centering
    \begin{minipage}{.7\textwidth}
        \centering
        \includegraphics[width=1\linewidth,height=8cm]{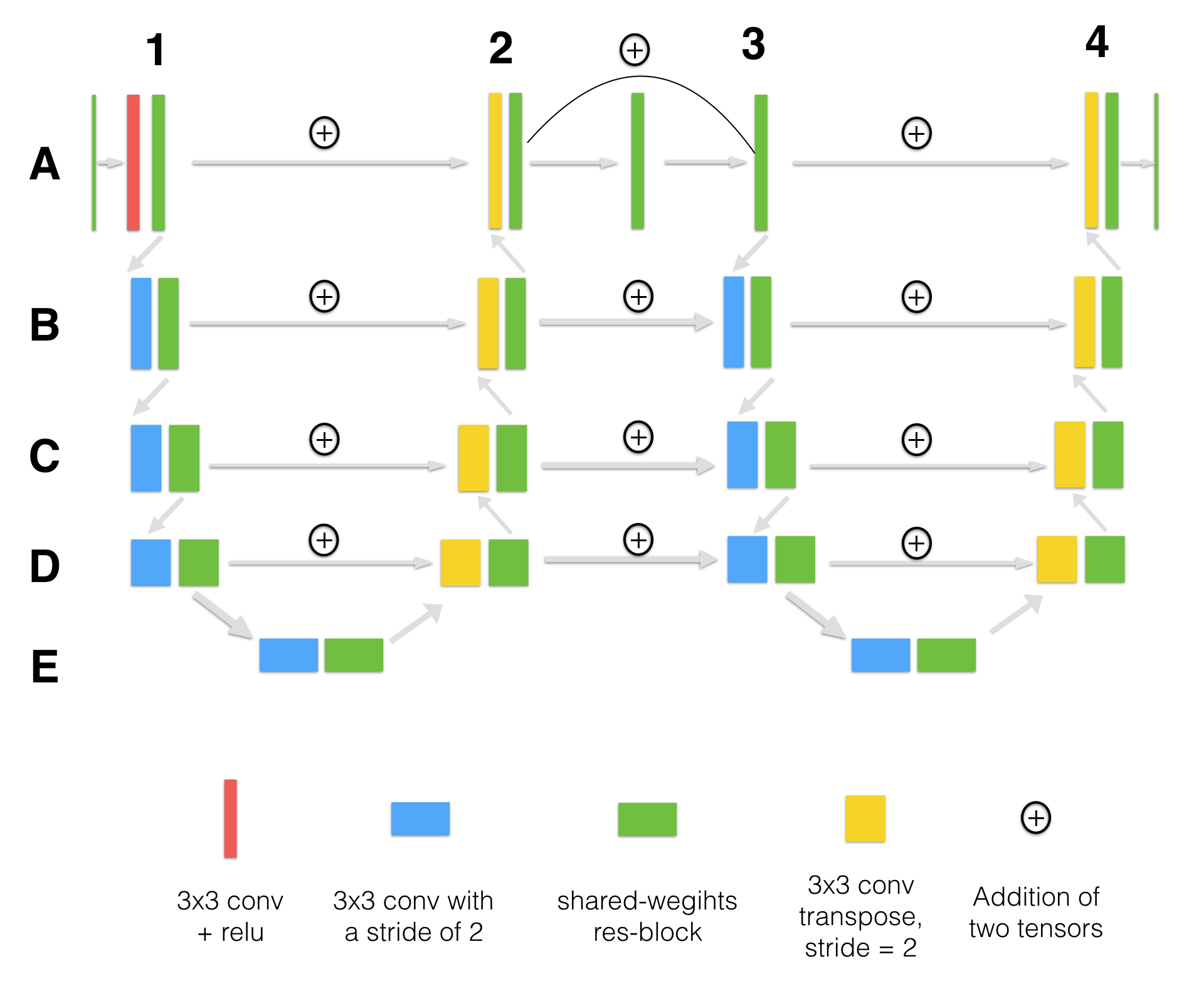}
    \end{minipage}%
    \begin{minipage}{0.3\textwidth}
        \centering
        \includegraphics[width=1\linewidth,height=8cm]{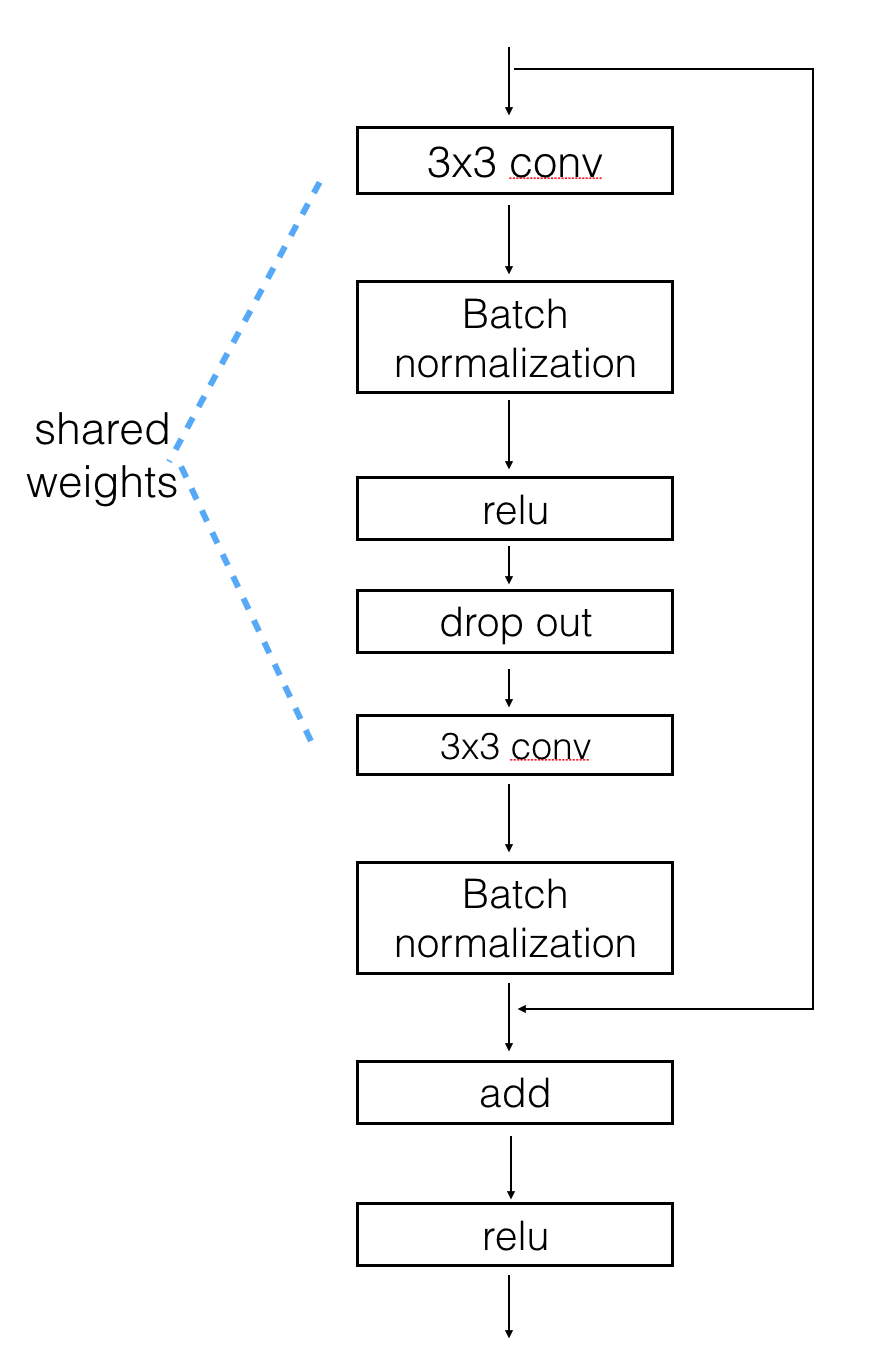}
    \end{minipage}
    \caption{
    \footnotesize{From left to right: Structure of LadderNet and shared-weights residual block.}
    }
    \label{laddernet}
\end{figure*} 
\vspace{-0.2cm}
\section{Methods}
\vspace{-0.2cm}
\subsection{LadderNet}
U-Net and its variants in the literature all have an encoder-decoder structure. However, the number of paths for information flow in U-Net is limited. We propose LadderNet, a multi-branch convolutional neural network for semantic segmentation as shown in Fig. \ref{laddernet}, which has more paths of information flow. Features in different spatial scales are named with letters A to E, and columns are named with numbers 1 to 4. We name column 1 and 3 as encoder branches, and name column 2 and 4 as decoder branches. We use convolution with a stride of 2 to go from small-receptive-field features to large-receptive-field features (e.g., A to B), and use transposed convolution with a stride of 2 to go from large-receptive-field features to large-receptive-field features (e.g., B to A). The number of channels is doubled from one level to the next level (e.g., A to B). 
\vspace{-0.1cm}
\subsection{Connection between LadderNet and U-Net}
LadderNet can be viewed as a chain of U-Nets. Columns 1 and 2 can be viewed as a U-Net, and Columns 3 and 4 can be viewed as another U-Net. Between two U-Nets, there are skip connections at levels A-D. Different from U-Net, where features from encoder branches are concatenated with features from decoder branches, we sum features from two branches. We demonstrate a LadderNet composed of 2 U-Nets here, but attach more U-Nets to form complicated network structures.
\par
LadderNet can also be viewed as an ensemble of multiple FCNs. Veit et al. proposed that ResNet behaves like an ensemble of shallow networks \cite{veit2016residual}, because the residual connections provide multiple paths of information flow. Similarly, LadderNet provides multiple paths of information flow, and we lists a few paths here as an example: (1) $A1 \rightarrow A2 \rightarrow A3 \rightarrow A4$, (2) $A1 \rightarrow A2 \rightarrow A3 \rightarrow B3 \rightarrow B4 \rightarrow A4$, (3) $A1 \rightarrow B1 \rightarrow B2 \rightarrow B3 \rightarrow B4 \rightarrow A4$, (4) $A1 \rightarrow B1 \rightarrow C1 \rightarrow D1 \rightarrow E1 \rightarrow D2 \rightarrow C2 \rightarrow B2 \rightarrow A2 \rightarrow A3 \rightarrow A 4$. Each path can be viewed as a variant of FCN. The total number of paths grows exponentially with the number of encoder-decoder pairs and the number of spatial levels (e.g., A to E in Fig. \ref{laddernet}). Therefore, LadderNet has the potential to capture more complicated features and produce a higher accuracy.
\vspace{-0.1cm}
\subsection{Shared-weights residual block}
More encoder-decoder branches will increase the number of parameters and the difficulty of training. To solve this problem, we propose shared-weights residual blocks as shown in Fig. \ref{laddernet}.  Different from a standard residual convolutional block proposed by He \cite{he2016deep}, the two convolutional layers in the same block share the same weights. Similar to the recurrent convolutional neural network (RCNN) \cite{alom2017inception}, the two convolutional layers in the same block can be viewed as one recurrent layer, except that the two batch normalization layers are different. A drop-out layer is added between two convolutional layers to avoid overfitting. The shared-weights residual block combines the strength of skip connection, recurrent convolution and drop-out regularization, and has much fewer parameters that a standard residual block. 

\section{Experiments}
\vspace{-0.1cm}
\subsection{Datasets}
We evaluated the proposed LadderNet on two popular datasets for retina blood vessel segmentation: the DRIVE dataset and the CHASE\_DB1 dataset. The DRIVE dataset consists of 40 color images of the retina, 20 of which were used for training and the remaining 20 images for testing. Each image has $565 \times 584$ pixels. To increase the number of training samples, we randomly sampled 190,000 patches of size $48 \times 48$ from the training images, and used $10\%$ of the training samples as validation data. 
\par 
The CHASE\_DB1 dataset was collected from both left and right eyes of 14 school children. It has 28 color images of the retina, 20 of which were used for training and the other 8 images (from 4 children) for testing. The size of each image is $996 \times 960$. We randomly sampled 760,000 patches of size $48 \times 48$ from the training images, and used $10\%$ of the training samples as validation. 
\par
All patches were converted to gray-scale for experiments. Field of view (FOV) is provided for the DRIVE dataset but not the CHASE\_DB1 dataset. We applied similar techniques in \cite{soares2006retinal} to generate FOV masks, and sampled patches over the entire image including regions outside of the FOV. 

\subsection{Training of the neural network}
We chose a LadderNet with 5 levels (A-E) and a drop-out rate of 0.25, and set the number of channels of the first level (level A) as 10, resulting in a LadderNet with 1.5M parameters. We used cross-entropy loss for semantic segmentation, and applied Adam optimizer with default parameters and a batch size of 1024. We used "reduce learning rate on plateau" strategy, and set the learning rate as 0.01, 0.001, 0.0001 on epochs 0, 20 and 150 respectively, and set the total learning epochs as 250. 

\subsection{Evaluation approaches}
We used several metrics to evaluate the performance of LadderNet, including accuracy (AC), sensitivity (SE), specificity (SP) and F1-score. First we calculated True Positive (TP), True Negative (TN), False Positive (FP) and False Negative (FN). Different metrics are calculated as follows:
\begin{equation}
AC = \frac{TP+TN}{TP+TN+FP+FN}
\end{equation}
\begin{equation}
SE=\frac{TP}{TP+FN}
\end{equation}
\begin{equation}
SP=\frac{TN}{TN+FP}
\end{equation}
The F1-score is calculated as follows:
\begin{equation}
Precision = \frac{TP}{TP+FP}
\end{equation}
\begin{equation}
Recall = \frac{TP}{TP+FN}
\end{equation}
\begin{equation}
F1 = 2 \times \frac{Precision \times Recall}{Precision + Recall}
\end{equation}
To further evaluate the performance of different neural networks, we calculated the receiver operating characteristics (ROC) curve and the are under curve (AUC). 
\section{Results}

\begin{figure}[!htb]
    \centering
    \begin{minipage}{.25\textwidth}
        \centering
        \includegraphics[width=1\linewidth]{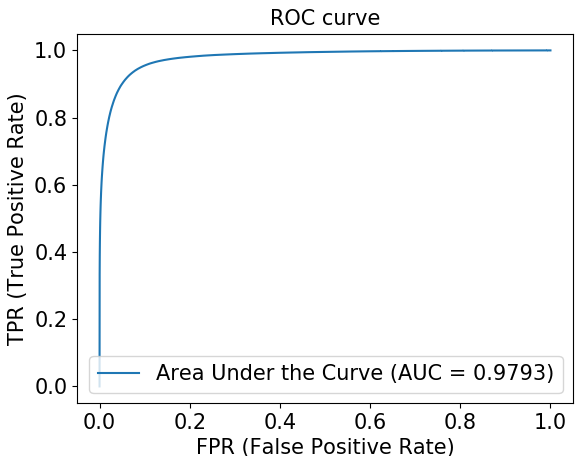}
    \end{minipage}%
    \begin{minipage}{0.25\textwidth}
        \centering
        \includegraphics[width=1\linewidth]{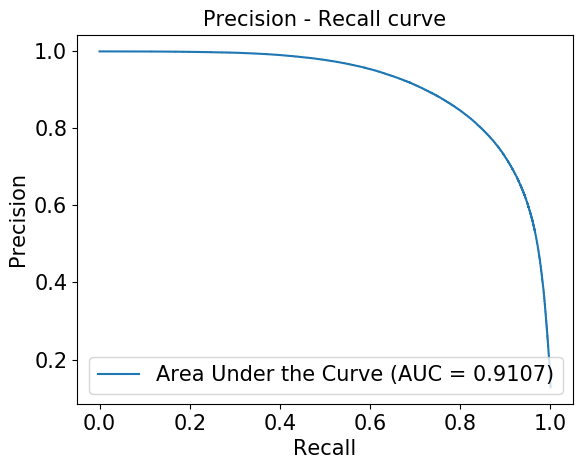}
    \end{minipage}
    \caption{
    \footnotesize{From left to right: Precision-recall curve and ROC curve on DRIVE dataset.}
    }
    \label{DRIVEROC}
\end{figure} 

\begin{figure}[!htb]
    \centering
    \begin{minipage}{0.14\textwidth}
        \centering
        \includegraphics[width=1\linewidth]{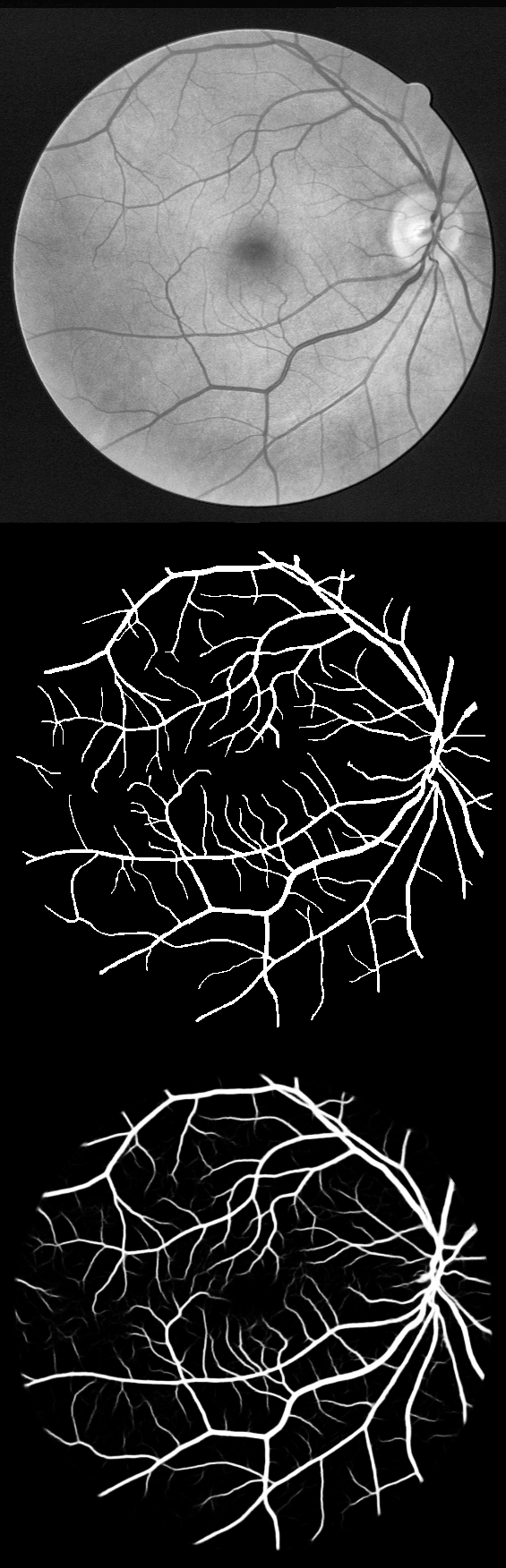}
    \end{minipage}%
    \begin{minipage}{0.14\textwidth}
        \centering
        \includegraphics[width=1\linewidth]{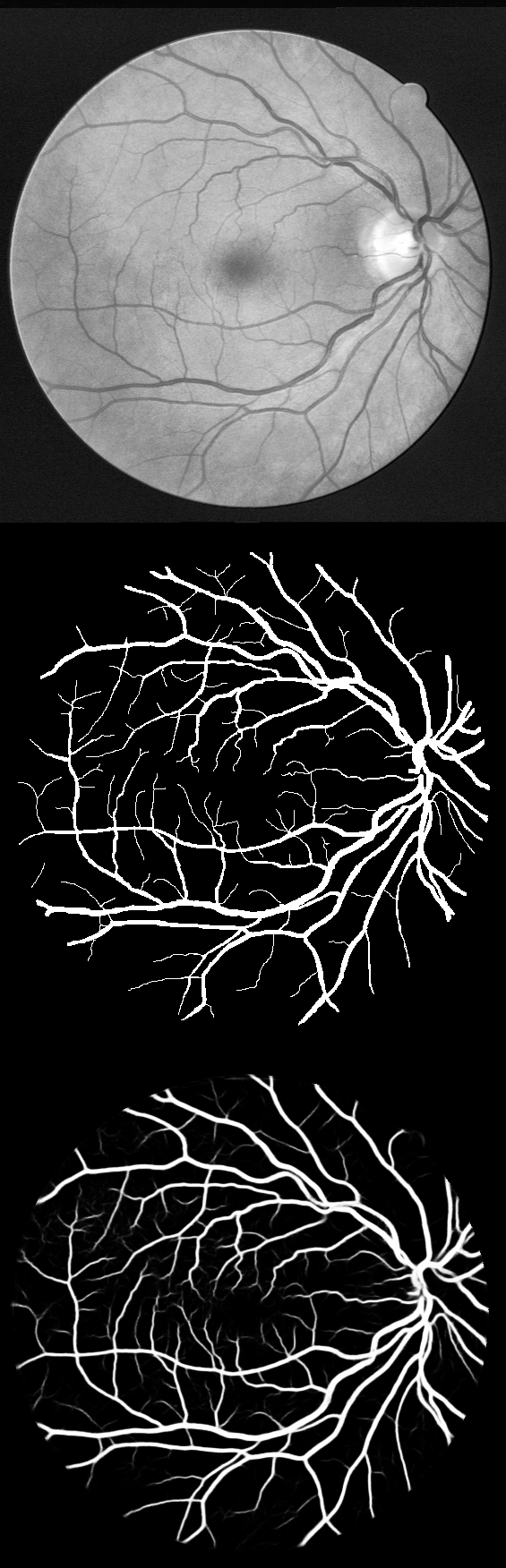}
    \end{minipage}%
    \begin{minipage}{0.14\textwidth}
        \centering
        \includegraphics[width=1\linewidth]{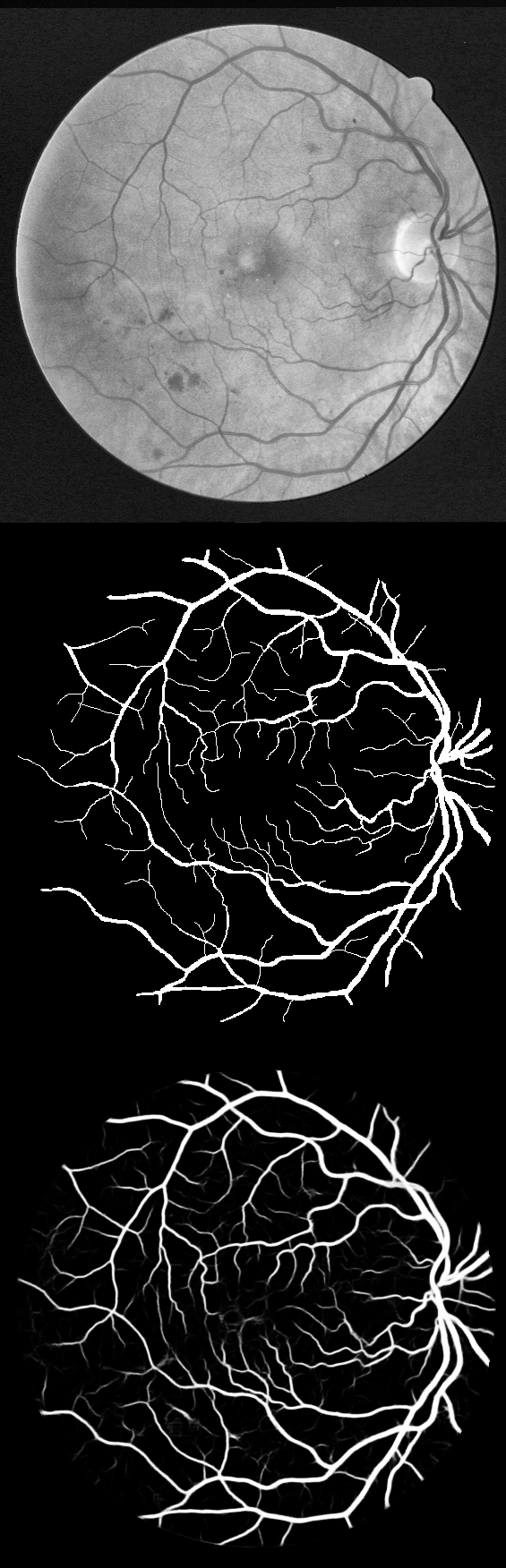}
    \end{minipage}%
    \caption{
    \footnotesize{Test results on DRIVE dataset. From Top to bottom: input image, ground truth and predictions.}
    }
    \label{DRIVEImage}
\end{figure}

\begin{figure}[!htb]
    \centering
    \begin{minipage}{.25\textwidth}
        \centering
        \includegraphics[width=1\linewidth]{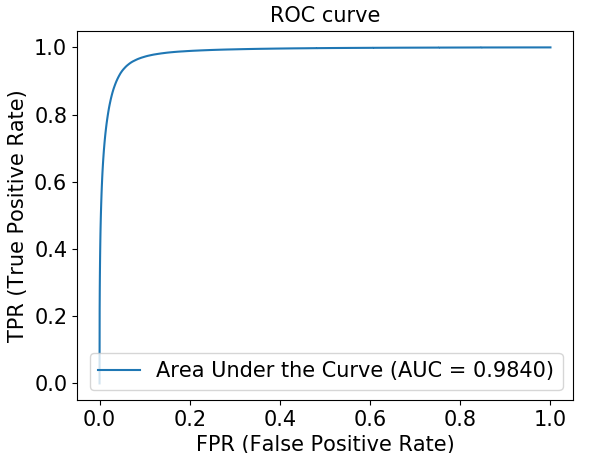}
    \end{minipage}%
    \begin{minipage}{0.25\textwidth}
        \centering
        \includegraphics[width=1\linewidth]{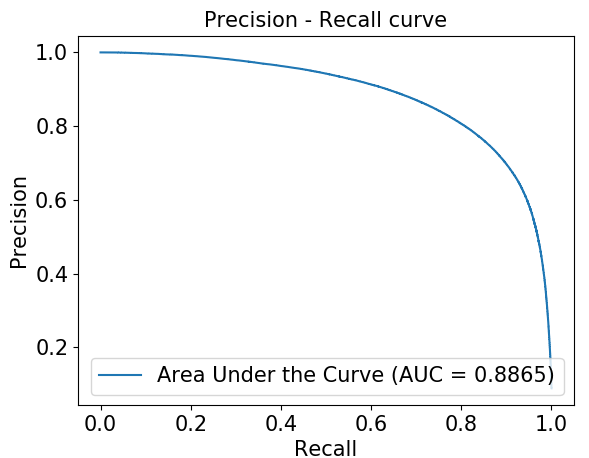}
    \end{minipage}
    \caption{
    \footnotesize{From left to right: Precision-recall curve and ROC curve on CHASE\_DB1 dataset.}
    }
    \label{CHASEROC}
\end{figure} 

\begin{figure}[!htb]
    \centering
    \begin{minipage}{0.14\textwidth}
        \centering
        \includegraphics[width=1\linewidth]{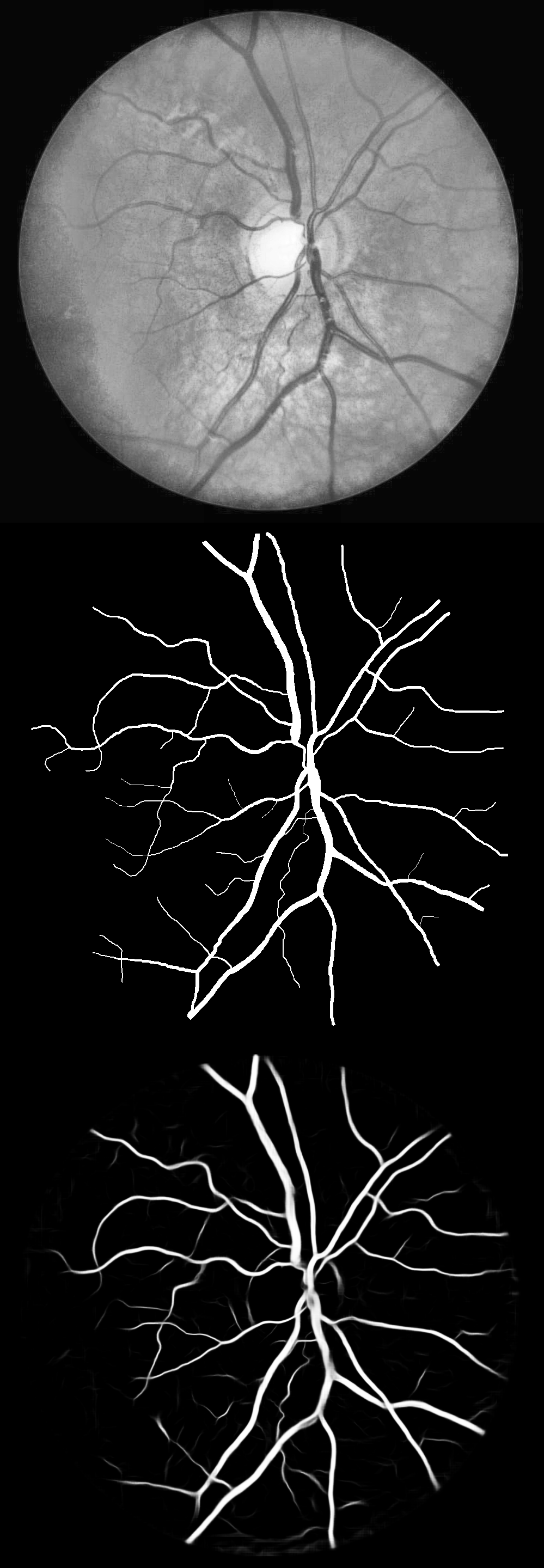}
    \end{minipage}%
    \begin{minipage}{0.14\textwidth}
        \centering
        \includegraphics[width=1\linewidth]{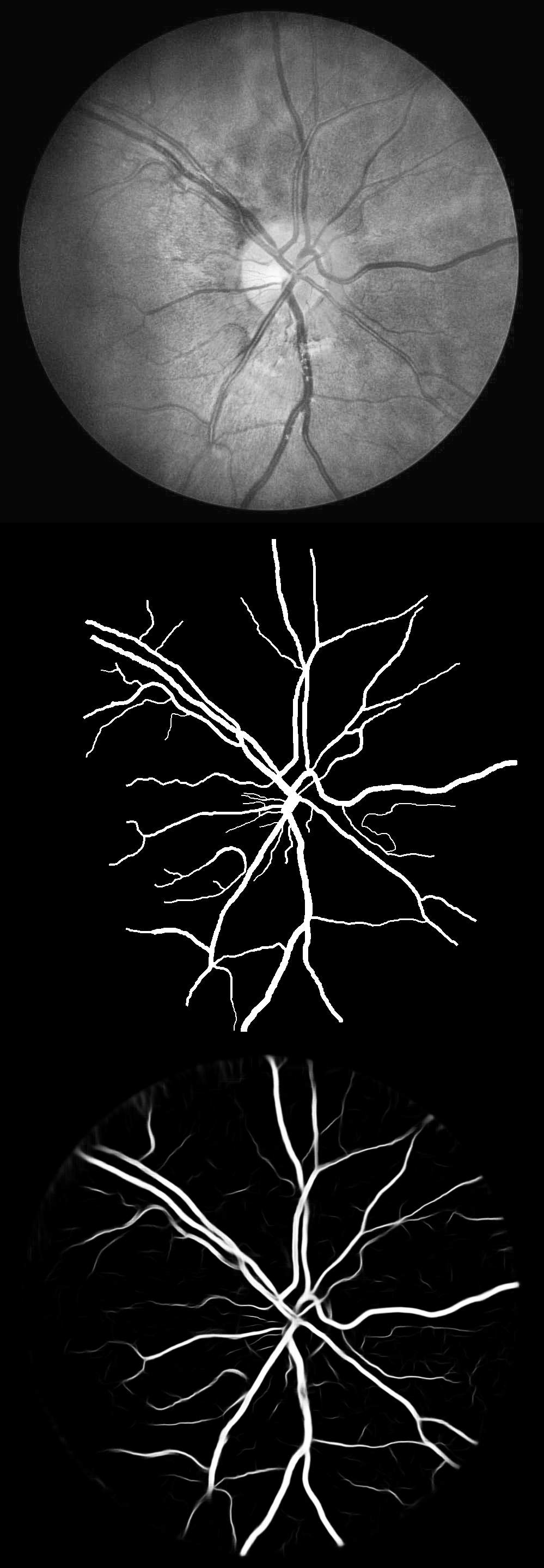}
    \end{minipage}%
    \begin{minipage}{0.14\textwidth}
        \centering
        \includegraphics[width=1\linewidth]{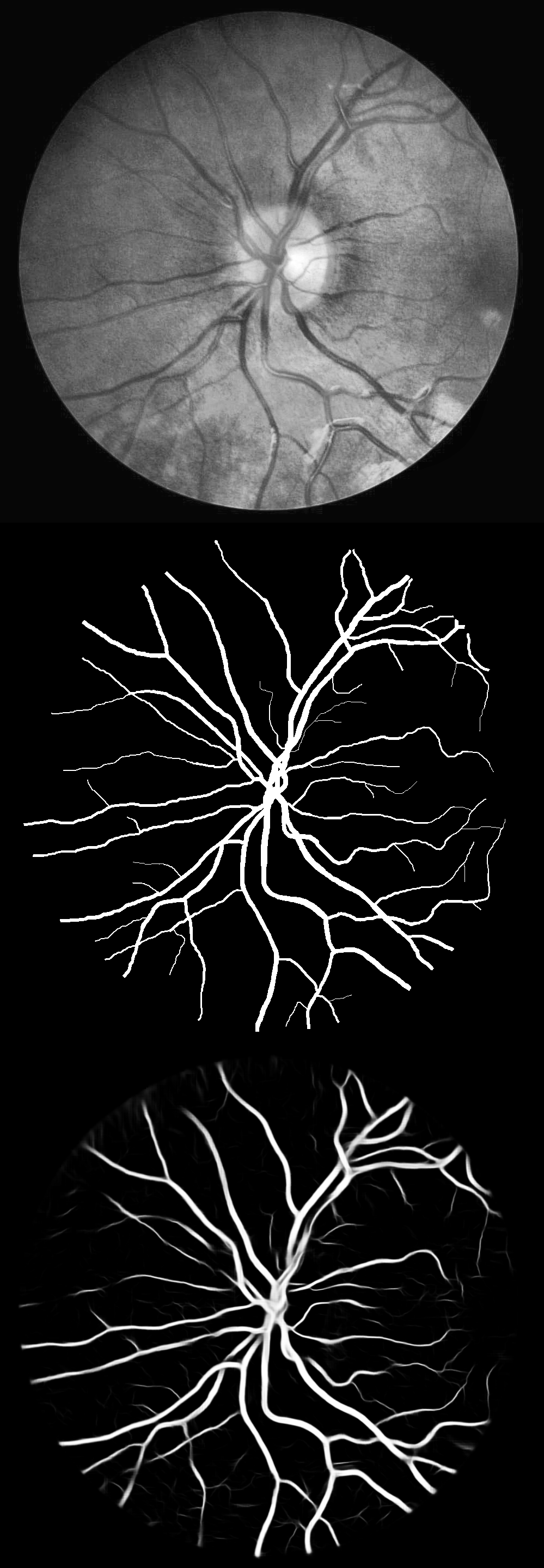}
    \end{minipage}%
    \caption{
    \footnotesize{Test results on CHASE\_DB1 dataset. From Top to bottom: input image, ground truth and predictions.}
    }
    \label{ChaseImage}
\end{figure} 
\begin{table}[]
\scalebox{0.62}{
\begin{tabular}{@{}|l|l|l|l|l|l|l|l|@{}}
\toprule
Dataset                      & Methods            & Year & F1-score        & SE              & SP              & AC              & AUC             \\ \midrule
\multirow{11}{*}{DRIVE}      & Roychowdhury \cite{roychowdhury2015blood} & 2016 & - & 0.7250 & \textbf{0.9830} & 0.9520 & 0.9620 \\ \cmidrule(l){2-8} 
                             & Qiaoliang Li \cite{li2016cross}      & 2016 & -               & 0.7569          & 0.9816          & 0.9527          & 0.9738          \\ \cmidrule(l){2-8} 
                             & U-Net    \cite{alom2018recurrent}           & 2018 & 0.8142          & 0.7537          & 0.9820          & 0.9531          & 0.9755          \\ \cmidrule(l){2-8} 
                             & Residual UNet  \cite{alom2018recurrent}     & 2018 & 0.8149          & 0.7726          & 0.9820          & 0.9553          & 0.9779          \\ \cmidrule(l){2-8} 
                             & Recurrent UNet  \cite{alom2018recurrent}   & 2018 & 0.8155          & 0.7751          & 0.9816          & 0.9556          & 0.9782          \\ \cmidrule(l){2-8} 
                             & R2U-Net       \cite{alom2018recurrent}     & 2018 & 0.8171          & 0.7792          & 0.9813          & 0.9556          & 0.9784          \\ \cmidrule(l){2-8} 
                             & \textbf{LadderNet} & 2018 & \textbf{0.8202} & \textbf{0.7856} & 0.9810          & \textbf{0.9561} & \textbf{0.9793} \\ \midrule
\multirow{11}{*}{CHASE\_DB1} & Roychowdhury \cite{roychowdhury2015blood} & 2016 & - & 0.7201 & 0.9824 & 0.9850 & 0.9532 \\ \cmidrule(l){2-8} 
                             & Qiaoliang Li  \cite{li2016cross}     & 2016 & -               & 0.7507          & 0.9793          & 0.9581          & 0.9793          \\ \cmidrule(l){2-8} 
                             & U-Net        \cite{alom2018recurrent}      & 2018 & 0.7783          & \textbf{0.8288} & 0.9701          & 0.9578          & 0.9772          \\ \cmidrule(l){2-8} 
                             & Residual U-Net \cite{alom2018recurrent}     & 2018 & 0.7800          & 0.7726          & 0.9820          & 0.9553          & 0.9779          \\ \cmidrule(l){2-8} 
                             & Recurrent U-Net  \cite{alom2018recurrent}  & 2018 & 0.7810          & 0.7459          & \textbf{0.9836} & 0.9622          & 0.9803          \\ \cmidrule(l){2-8} 
                             & R2U-Net   \cite{alom2018recurrent}         & 2018 & 0.7928          & 0.7756          & 0.9820          & 0.9634          & 0.9815          \\ \cmidrule(l){2-8} 
                             & \textbf{LadderNet}          & 2018 & \textbf{0.8031} & 0.7978          & 0.9818          & \textbf{0.9656} & \textbf{0.9839} \\ \bottomrule
\end{tabular}}

\caption{Results of LadderNet and other methods on DRIVE and CHASE\_DB1 datasets.}
\label{Table}
\end{table}

Results of LadderNet are shown in Figs. 2-5. LadderNet generates predictions that are visually very close to the ground truth, and the areas under the ROC curves are above 0.97 for both tasks, and the areas under the precision-recall curves are above 0.88 for both tasks.
\par
Table \ref{Table} demonstrates the quantitative results of different methods. LadderNet generates the highest F1-score, accuracy and AUC for both tasks. LadderNet also generates high SE and SP on two tasks. It's easy to generate a high SE or SP by predicting image pixels towards one category more easily than the other; in the extreme case, predicting the entire image as blood vessel will generate a $SE=1.0$ but $SP=0.0$. SE and SP focus more on one category than the other, while other metrics such as AC, AUC and F1-score evaluates the performance of a model based on performance on two categories; therefore, a higher AC, AUC and F1-score is more convincing than a higher SE or SP. LadderNet achieves the highest AU, AUC and F1-score in both tasks, therefore it performs the best compared to previous models.

\section{Discussion and conclusion}
We propose LadderNet for semantic segmentation in this paper. Compared to U-Net, LadderNet has more encoder-decoder pairs. The skip connections enable LadderNet to have multiple paths for information flow, and the number of paths increases exponentially with the number of encoder-decoder pairs. Another innovation is the shared-weights residual block, which combines the strengths of residual connection, recurrent convolutional layer and drop-out regularization. The shared-weights design also greatly reduces the number of parameters. Our LadderNet has a superior performance over previous methods in the literature on two public datasets, and improves the performance on the key problem to automatic retinal disease detection. LadderNet can also be used in other semantic segmentation tasks such as tumor segmentation or brain lesion detection.
\bibliographystyle{IEEEbib}
\small{
\bibliography{strings,refs}
}
\end{document}